\documentclass[cameraready]{Interspeech}

\usepackage{booktabs}
\usepackage{graphicx}
\usepackage{multirow}
\usepackage{amsmath}
\usepackage{siunitx}
\usepackage[table]{xcolor}

\usepackage[utf8]{inputenc}
\usepackage[T1]{fontenc}
\DeclareUnicodeCharacter{1E6D}{\d{t}}
\DeclareUnicodeCharacter{1E6C}{\d{T}}

\newcommand{\tdot}{\d{t}}

\title{Towards English–Mizo Legal Machine Translation: Corpus Creation, Analysis, and Benchmarks}

\author[orcid=0000-0001-5794-844X, correspondingauthor]{Telem Joyson}{Singh}
\author[]{Sanabam}{Boyshana}
\author[orcid=0000-0002-9051-1255]{Sanasam Ranbir}{Singh}
\author[orcid=0000-0003-0484-2144]{Priyankoo}{Sarmah}

\address{
    Indian Institute of Technology Guwahati, Assam, India
}

\email{tjoyson@iitg.ac.in, boyshanasanabam@gmail.com, ranbir@iitg.ac.in, priyankoo@iitg.ac.in}


\begin{document}

\maketitle

\begin{abstract}
Indian court proceedings are held in English, so speakers of languages with limited translation technology have limited access to them. We present the first English–Mizo parallel dataset of Indian court judgments. It contains 13,168 sentence pairs from the Gauhati High Court and the Supreme Court of India. Professional native Mizo translators translated complete documents, sentence pairs were manually aligned, and two independent reviewers checked the results. Our study finds that English words are borrowed in a very selective way: official and legal terms are almost always kept in English, but words for everyday things like actions, situations, and results are usually expressed in native Mizo. Mizo diacritics are also used inconsistently across the corpus. When we fine-tune the NLLB-200 1.3B model on our training data, it scores 27.47 BLEU on a balanced legal test set of 1,000 sentence pairs—an improvement of 7.16 BLEU over the base model. However, it still uses English words about three times more often than a human translator does.
\end{abstract}

\section{Introduction}

Access to justice depends on understanding the language of the courtroom. In India, Article~348 of the Constitution requires that proceedings in the Supreme Court and the High Courts be conducted in English, so the official text of judgments is in English even when the parties cannot read it. The Supreme Court and the e-Courts project have launched translation initiatives, but these target the languages listed in the Eighth Schedule \cite{eCourtsPhase3Vision,SCI-AnnualReport2020}. Languages outside the Schedule receive little institutional support, and their communities are left to rely on informal, unverified oral summaries of the decisions that affect them.

Mizo (also called Lushai) is one such language. It is a Tibeto-Burman language spoken by about \num{830000} people (2011 Census of India \cite{CensusIndia2011}) and is an official language of Mizoram alongside English. Because it is not in the Eighth Schedule, it is usually excluded from the major Indic datasets and models~\cite{samanantar2022,indictrans2}. This gap is especially serious in the legal domain. Mizoram is under the Gauhati High Court, where the rulings are written only in English. As a result, Mizo speakers have almost no tools to help them read and understand court decisions themselves.

The main challenge is the domain, not just the volume of data. Most existing Mizo parallel texts are from religious writings or informal web content, which are very different from legal writing in length, sentence structure, citation style, and vocabulary. Models developed on such data transfer poorly to legal text, and until now there has been no legal test set to measure how large that gap is. As a result, progress has been less constrained by modeling capacity than by the lack of a reliable evaluation resource.

We address this with the first parallel corpus of Indian legal judgments for Mizo: \num{13168} sentence pairs, \num{286590} English and \num{291751} Mizo tokens, drawn from civil, criminal and miscellaneous cases of the Gauhati High Court and the Supreme Court of India. Professional native Mizo translators with degrees in law, English, or Mizo translated entire documents, so that references and terminology remain consistent within a judgment. We then aligned the texts at the sentence level using embedding-based candidate generation followed by manual correction, and validated the result with two independent annotators (Cohen's $\kappa = 0.90$).

Beyond releasing the data, we analyze its linguistic properties. Because Mizo has no indigenous common-law vocabulary, translating Indian legal English into Mizo involves systematic borrowing: \num{6612} word types occur on both sides of the corpus, and \SI{55.30}{\percent} of Mizo tokens are strings that also appear in the English vocabulary. The borrowing is selective. Names of institutions and statutory instruments---\textit{court}, \textit{FIR}, \textit{IPC}, and \textit{affidavit}—are kept in English almost all the time (close to \SI{100}{\percent}). In contrast, words referring to people, legal processes, and outcomes---\textit{accused} (\SI{8.5}{\percent}) and \textit{conviction} (\SI{5.0}{\percent})—are usually replaced with native Mizo words. We link this pattern to \emph{referential rigidity}: the English form is kept exactly in those cases where replacing it could create ambiguity. We also observe substantial orthographic variation on the Mizo side. Circumflex tone marking appears only in \SI{2.70}{\percent} of Mizo segments, \num{159} types are attested both with and without it, and the \textit{t}/\textit{\tdot} contrast is largely unrealized--variation that affects tokenization and comparability between systems.

Finally, we create baseline systems by fine-tuning the NLLB-200 \num{1.3}B model \cite{nllb2022}—one of the rare open-source, large-scale translation models with explicit Mizo—using the \num{11522}-pair training set. On a stratified \num{1000}-pair held-out legal test , the model scores \num{27.47} BLEU, \num{50.39} chrF++, and \num{65.68} TER. Compared to the original, non-adapted model, this is an improvement of $+\num{7.16}$ BLEU, $+\num{7.83}$ chrF++, and a reduction of $-\num{12.13}$ TER. The error analysis shows that most of the remaining mistakes happen when the system has to decide whether to borrow English words. For legal terms, it keeps the English word in \SI{21.7}{\percent} of the cases where the human translation uses a native Mizo word, but only \SI{8.3}{\percent} the other way around. In other words, it borrows English terms about three times more often than the human translator.

\noindent Our contributions are as follows.
\begin{itemize}
  \item We release the first English--Mizo parallel corpus in the legal domain: \num{13168} professionally translated, manually aligned, and independently validated sentence pairs from Indian court judgments.
  \item We analyze the corpus quantitatively, including a retention-rate study of legal terminology that shows a systematic borrowing pattern, and an audit of Mizo orthography that identifies diacritic normalization as an open preprocessing question for Mizo NLP.
  \item We present both fine-tuned models and baseline systems on a stratified held-out legal test set, including an error analysis by stratum, thereby providing reference points for future research on Mizo legal translation.
\end{itemize}

\section{Background}

\subsection{The Mizo language}
\label{ssec:mizo-language}

Mizo is spoken mainly in Mizoram, with communities in neighboring northeast Indian states, Myanmar's Chin State, and Bangladesh's Chittagong Hill Tracts \cite{chhangte1989,chhangte1993}. The standard written variety is based on the Duhlian dialect.

Three properties matter for our corpus. First, Mizo is agglutinative, with extensive particles and derivational affixation, so a root appears on many surfaces; this is shown in the vocabulary and hapax counts in Section~\ref{sec:corpus-analysis}. Second, Mizo is tonal, but this tone is only partly written. Third, its writing system is based on the Latin alphabet and was created by Christian missionaries in the late nineteenth century. It includes two features that English does not have: the special letter \textit{\tdot} (\textsc{latin small letter t with a dot below}), which represents a different sound from plain \textit{t}, and vowels with a circumflex accent that show both tone and vowel length. Both are optional in everyday writing, so running text stays often under-specified relative to the prescriptive standard; we quantify this in Section~\ref{ssec:orthography}.

That missionary history similarly shapes the corpus. For decades, Bible translation was the main literate use of written Mizo, so religious texts dominate what is digitally available, while technical and administrative writing is scarce. Common-law concepts entered Mizo through English-medium administration and were never lexicalized as a coherent native system, so translating judicial prose depends on the borrowing behavior analyzed in Section~\ref{ssec:borrowing}.

\subsection{Language technology for Mizo}
\label{ssec:mizo-nlp}

Indian-language NLP has advanced mainly through corpora and models built for twenty-two scheduled languages \cite{samanantar2022,indictrans2,telem-joyson-etal-2023-big}. Mizo is not a scheduled language, so most of these resources exclude it.

Mizo has entered modern multilingual NLP largely through massively multilingual projects. It appears as \texttt{lus\_Latn} in the NLLB-200 models and the FLORES benchmark \cite{nllb2022,nllb2024nature}, which makes NLLB-200 a natural starting point and motivates our choice of base model in Section~\ref{sec:experimental-setup}. Coverage does not guaranty quality, however: the Mizo training data in such systems is largely religious and crawled web text, and the evaluation sets are general-domain. Mizo-focused work has focused on building general-purpose parallel data and supervised and unsupervised NMT for the pair \cite{lalrempuii2021improved,khenglawt2022wildre,lalrempuii2023unsupervised}; to our knowledge, no resource in the legal-domain has been available.

\subsection{Legal-domain translation and corpora}
\label{ssec:legal-mt}

Legal text remains a hard case for machine translation. Judgments use long, deeply nested sentences, fixed citation and case-title patterns, formal honorific language, and terms whose near-synonyms are not interchangeable: translating \textit{accused} as \textit{convict}, or \textit{petitioner} as \textit{appellant} changes the legal meaning. Here, correct terminology matters more than fluent language, and mistakes can harm the litigant.

In Europe, large multilingual legal corpora produced by institutional translation-- parliamentary proceedings, statutes, and case law \cite{koehn2005europarl,steinberger2006jrcacquis,steinberger2012dgttm}---have supported extensive work on legal MT and terminology. In India, legal NLP has grown to include judgment summarisation, rhetorical role labeling, as well as pretrained legal language models, supported by parallel datasets for several Indian languages \cite{assamlegaltrans,parikh2021lawsum,mahapatra2025milpac}. These resources stop at the scheduled languages, so Mizo is absent. Our corpus extends this line of work to a non-scheduled language.

\section{Corpus Construction}
\label{sec:corpus-construction}

The corpus is made up of English-language judgments from the Gauhati High Court and the Supreme Court of India, including civil, criminal, and other types of cases. All documents come from official court records that are already publicly available, so we did not apply any extra anonymization beyond how the judgments are originally published.

\noindent\textbf{Document translation.}
The English judgments were translated into Mizo by a team of professional translators, all native Mizo speakers holding degrees in law, English, or Mizo. Translation was performed at the document level rather than sentence by sentence, so that translators could use the larger context to resolve ambiguity and keep terminology consistent within a judgment. Then each translated document was peer-reviewed by a second translator.

\noindent\textbf{Sentence alignment.}
After translation and review, we aligned the documents at sentence level. Both sides were first segmented into sentences, using spaCy \cite{spacy} for English and NLTK \cite{bird2009nltk} for Mizo. We then used an aligner that relies on multilingual sentence embeddings to generate possible sentence matches, and we ranked these candidates using cosine similarity. Since legal sentences are often long and complicated, we did not rely on automation alone: human annotators went through the suggested segments, fixed one-to-many and many-to-one mappings and made sure each matched pair had the same meaning.

\noindent\textbf{Validation.}
Two native Mizo speakers independently reviewed every pair to confirm that the two sentences were translations of each other. Inter-annotator agreement was high (Cohen's $\kappa = 0.90$ \cite{cohen1960kappa}). Disagreements were referred to a third translator and resolved by a majority vote.

\section{Corpus Analysis}
\label{sec:corpus-analysis}

This section describes the corpus in three dimensions: basic word statistics, how legal terms are borrowed and kept, and how consistently Mizo spelling is used. For all of this, we count only alphabetic tokens using Unicode-aware tokenization, and we exclude punctuation and standalone numbers.

\subsection{Descriptive and lexical statistics}
\label{ssec:descriptive}

\begin{table}[t]
\centering
\small
\caption{Descriptive statistics for the English--Mizo corpus.}
\label{tab:descriptive}
\begin{tabular}{lrr}
\toprule
\textbf{Statistic} & \textbf{English} & \textbf{Mizo} \\
\midrule
Sentence pairs            & \num{13168}   & \num{13168} \\
Running tokens            & \num{286590}  & \num{291751} \\
Vocabulary (types)        & \num{10317}   & \num{11996} \\
Type--token ratio         & 0.036         & 0.041 \\
Hapax legomena            & \num{2476}    & \num{4253} \\
\quad as \% of types      & 24.00         & 35.45 \\
Characters                & \num{1659854} & \num{1573253} \\
Characters per word       & 4.66          & 4.27 \\
\midrule
Mean sentence length      & 21.76         & 22.16 \\
Median sentence length    & 16.0          & 17.0 \\
Standard deviation        & 17.44         & 17.09 \\
5th--95th percentile      & 5--56         & 5--55 \\
Maximum sentence length   & 223           & 186 \\
\bottomrule
\end{tabular}
\end{table}

Table~\ref{tab:descriptive} summarizes the corpus at the segment and token level. It contains \num{13168} sentence pairs, \num{286590} English, and \num{291751} Mizo running tokens.

With almost the same number of tokens, the Mizo side has \SI{16}{\percent} more types than the English side, and \SI{35.45}{\percent} of its types occur only once, compared to \SI{24.00}{\percent} in English. Two factors contribute. First, Mizo derivational and particle morphology produces a larger inventory of distinct surface forms. Second, the corpus has a long tail of Mizo personal and place names that appear in cause titles and are transferred unchanged to both sides. The lower character-per-word figure for Mizo indicates its predominantly CV and CVC syllable structure.

\begin{table}[t]
\centering
\small
\caption{Cumulative token coverage by vocabulary rank (\%).}
\label{tab:coverage}
\begin{tabular}{lrr}
\toprule
\textbf{Top-$k$ types} & \textbf{English} & \textbf{Mizo} \\
\midrule
\num{100}   & 52.73 & 51.55 \\
\num{500}   & 73.98 & 74.96 \\
\num{1000}  & 82.09 & 83.11 \\
\num{5000}  & 96.69 & 96.34 \\
\bottomrule
\end{tabular}
\end{table}

\begin{figure}[!t]
    \centering
    \includegraphics[width=0.99\linewidth]{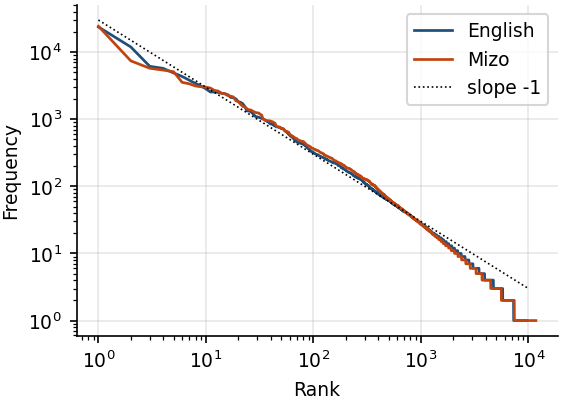}
    \caption{Zipf distribution of token frequencies.}
    \label{fig:zipf}
\end{figure}

\begin{figure}[!t]
    \centering
    \includegraphics[width=0.99\linewidth]{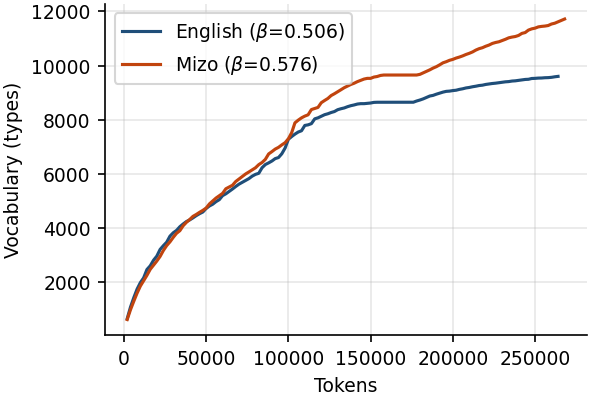}
    \caption{Vocabulary growth (Heaps' law).}
    \label{fig:heaps}
\end{figure}

Table~\ref{tab:coverage} reports cumulative token coverage by frequency rank. Beyond the first hundred types, the two languages are closely matched, with Mizo slightly lower in rank \num{5000}, consistent with its heavier type tail. Figures~\ref{fig:zipf} and~\ref{fig:heaps} show that both sides follow the expected Zipfian frequency distribution and Heaps' law vocabulary growth, with a higher Heaps exponent for Mizo ($\beta = 0.576$ vs.\ $0.506$). Mizo vocabulary keeps rising sharply with no clear plateau at 13,168 sentence pairs, indicating the legal corpus still lacks full lexical coverage and needs more data.

\subsection{Lexical borrowing and terminology retention}
\label{ssec:borrowing}

Mizo doesn’t have its own traditional legal terms, so when Indian legal English is translated into Mizo, new words almost always have to be borrowed. This subsection measures how often that happens, and we see this borrowing as the most distinctive language feature in the whole set of the corpus.

The aggregate borrowing is substantial (Table~\ref{tab:borrowing-summary}): \num{6612} word types occur on both sides, \SI{55.30}{\percent} of Mizo running tokens are strings that also appear in the English vocabulary, and the mean Jaccard token-level overlap per-pair is $0.4315$.

\begin{table}[t]
\centering
\small
\caption{Corpus-level lexical overlap (English--Mizo).}
\label{tab:borrowing-summary}
\begin{tabular}{lr}
\toprule
\textbf{Measure} & \textbf{Value} \\
\midrule
Shared word types & \num{6612} \\
Mizo tokens also in English vocab. & \SI{55.30}{\percent} \\
Mean token-level Jaccard (per-pair) & $0.4315$ \\
\bottomrule
\end{tabular}
\end{table}

To separate selective borrowing from wholesale copying, we compute a \emph{retention rate} for a hand-selected inventory of legal terms: the number of Mizo segments containing the term divided by the number of English segments containing it. Values near \SI{100}{\percent} indicate that the English form is carried over; low values indicate that the concept is lexicalized natively in Mizo. The results are in Table~\ref{tab:retention}.

\begin{table}[t]
\centering
\small
\caption{Retention of English legal terms in the Mizo text.
\textit{Ret.} is the proportion of Mizo segments to English segments
that include the term.}
\label{tab:retention}
\begin{tabular}{lrrr}
\toprule
\textbf{Term} & \textbf{En} & \textbf{Mz} & \textbf{Ret.\ (\%)} \\
\midrule
\multicolumn{4}{l}{\textit{High retention: institutions and instruments}} \\
CrPC          & 17& 17   & 100.0\\
FIR           & 46& 46   & 100.0\\
court         & \num{1616}& \num{1616} & 100.0\\
police        & 230& 230  & 100.0\\
IPC           & 47& 47   & 100.0\\
magistrate    & 223  & 222  & 99.6 \\
section       & 664  & 653  & 98.3 \\
respondent    & 606  & 574  & 94.7 \\
tribunal      & 16   & 15   & 93.8 \\
affidavit     & 185  & 173  & 93.5 \\
bail          & 219  & 201  & 91.8 \\
act           & 692  & 563  & 81.4 \\
advocate      & 396  & 309  & 78.0 \\
\midrule
\multicolumn{4}{l}{\textit{Intermediate}} \\
appeal        & 217  & 162  & 74.7 \\
appellant     & 285  & 211  & 74.0 \\
witness       & 80   & 54   & 67.5 \\
constitution  & 306  & 205  & 67.0 \\
petitioner    & 902  & 597  & 66.2 \\
counsel       & 664  & 323  & 48.6 \\
writ          & 232  & 112  & 48.3 \\
\midrule
\multicolumn{4}{l}{\textit{Low retention: statuses and reasoning}} \\
order         & 514  & 201  & 39.1 \\
suit          & 46   & 18   & 39.1 \\
evidence      & 197  & 73   & 37.1 \\
jurisdiction  & 86   & 24   & 27.9 \\
learned       & \num{1154} & 303 & 26.3 \\
judgment      & 280  & 66   & 23.6 \\
sentence      & 55   & 10   & 18.2 \\
decree        & 22   & 4    & 18.2 \\
accused       & 801  & 68   & 8.5 \\
conviction    & 40   & 2    & 5.0 \\
\bottomrule
\end{tabular}
\end{table}

The distribution is far from uniform, and the pattern is interpretable. Terms naming \emph{institutions, statutory instruments and fixed procedural labels}---\textit{court}, \textit{section}, \textit{FIR}, \textit{IPC}, \textit{CrPC}, \textit{affidavit}, \textit{magistrate}, \textit{bail}---are retained at rates approaching \SI{100}{\percent}. Terms naming \emph{statuses, acts and inferential outcomes}---\textit{accused} (\SI{8.5}{\percent}), \textit{conviction} (\SI{5.0}{\percent}), \textit{judgment} (\SI{23.6}{\percent}), \textit{evidence} (\SI{37.1}{\percent})---are overwhelmingly rendered with native Mizo words.

We see this pattern as measuring how strongly a term is tied to a specific reference. When a term refers to something whose identity is fixed by law or institutional designation and mistranslating it would create referential ambiguity, the English form is preserved. Where a term describes a condition, a process, or a conclusion that can be expressed compositionally, Mizo lexicalizes it: \textit{thubuai} for a case or dispute, \textit{dan} for law or act, \textit{hriattirna} for notice, \textit{thuneihna} for jurisdiction or authority. The rate for \textit{learned} (\SI{26.3}{\percent}) is a revealing minor case: the honorific register typical of Indian legal English is largely not carried in Mizo, so translation here involves register normalization and lexical substitution.

\subsection{Mizo orthographic consistency}
\label{ssec:orthography}

\begin{table}[t]
\centering
\small
\caption{Inventory of Mizo-specific orthographic characters on the Mizo side.}
\label{tab:charinv}
\begin{tabular}{llr}
\toprule
\textbf{Character} & \textbf{Name} & \textbf{Count} \\
\midrule
â & a with circumflex  & 294 \\
\tdot{} & t with dot below   & 282 \\
ê & e with circumflex  & 171 \\
û & u with circumflex  & 94  \\
î & i with circumflex  & 41  \\
\bottomrule
\end{tabular}
\end{table}

\begin{table}[t]
\centering
\small
\caption{Most frequent Mizo types attested with inconsistent circumflex
marking. Frequency is the combined count of both variants.}
\label{tab:diacritic}
\begin{tabular}{llr}
\toprule
\textbf{Base form} & \textbf{Attested variants} & \textbf{Freq.} \\
\midrule
chu     & chu / chû         & \num{5744} \\
te      & te / tê           & \num{2973} \\
tur     & tur / tûr         & \num{1858} \\
dan     & dan / dân         & \num{1416} \\
he      & he / hê           & \num{1257} \\
awm     & awm / âwm         & \num{1238} \\
tak     & tak / tâk         & 876 \\
dilna   & dilna / dîlna     & 666 \\
pek     & pek / pêk         & 657 \\
khan    & khan / khân       & 578 \\
\bottomrule
\end{tabular}
\end{table}

\begin{table}[!t]
\centering
\small
\caption{Realisation of the \textit{t}/\textit{\tdot} contrast for
selected lexemes.}
\label{tab:tdot}
\begin{tabular}{lrlr}
\toprule
\textbf{Plain} & \textbf{Freq.} & \textbf{Dotted} & \textbf{Freq.} \\
\midrule
tan   & 277 & \tdot{}an   & 12 \\
tha   & 111 & \tdot{}ha   & 27 \\
tul   & 109 & \tdot{}ul   & 16 \\
tawng & 13  & \tdot{}awng & 4  \\
\bottomrule
\end{tabular}
\end{table}

Written Mizo adds two features to the Latin alphabet: the letter \textit{\tdot}, which contrasts with \textit{t}, and circumflex vowels that mark tone and length. We assess how consistently the corpus uses both.
Table~\ref{tab:charinv} shows that both markers are rare. Circumflexes appear only in \SI{2.70}{\percent} of the Mizo segments and \SI{0.22}{\percent} of tokens, and \num{159} types occur both with and without them (Table~\ref{tab:diacritic}). The letter \textit{\tdot} is equally rare (\num{282} tokens in \num{52} types), and the counts in Table~\ref{tab:tdot} show that the \textit{t}/\textit{\tdot} contrast is largely not maintained.

We regard this as an open problem for Mizo NLP rather than a defect of the corpus. It is not yet clear whether models should be trained on text that keeps diacritics or on text where they are stripped, and the choice affects vocabulary size, segmentation cost, and the comparability of published results. We therefore release the corpus unnormalised, so that either convention can be applied downstream.

\section{Experimental Setup}
\label{sec:experimental-setup}

\begin{table}[t]
\centering
\small
\caption{Statistics of the training and test splits.}
\label{tab:corpus-stats}
\begin{tabular}{llrrrr}
\toprule
\textbf{Split} & \textbf{Lang.} & \textbf{Sent.} & \textbf{Tokens} & \textbf{Types} & \textbf{TTR} \\
\midrule
\multirow{2}{*}{Train} & English & \multirow{2}{*}{\num{11522}} & \num{245838} & \num{11929} & 0.049 \\
                       & Mizo &                              & \num{248039} & \num{13205} & 0.053 \\
\midrule
\multirow{2}{*}{Test}  & English & \multirow{2}{*}{\num{1000}}  & \num{21990}  & \num{3968}  & 0.180 \\
                       & Mizo &                              & \num{23150}  & \num{3669}  & 0.159 \\
\bottomrule
\end{tabular}
\end{table}

\subsection{Training data}
The training split contains \num{11522} sentence pairs from the corpus described in Section~\ref{sec:corpus-analysis}, with \num{245838} English and \num{248039} Mizo tokens (Table~\ref{tab:corpus-stats}). It is formed after the set of benchmark tests is removed, and any sentence with more than \SI{90}{\percent} words overlaps with a test sentence. As in the corpus as a whole, the Mizo side has the larger vocabulary (\num{13205} vs.\ \num{11929} types) and the higher TTR (0.053 vs.\ 0.049), consistent with the morphology and orthographic variation reported in Section~\ref{sec:corpus-analysis}.

\subsection{Test data}
\begin{figure}[t]
    \centering
    \includegraphics[width=0.95\linewidth]{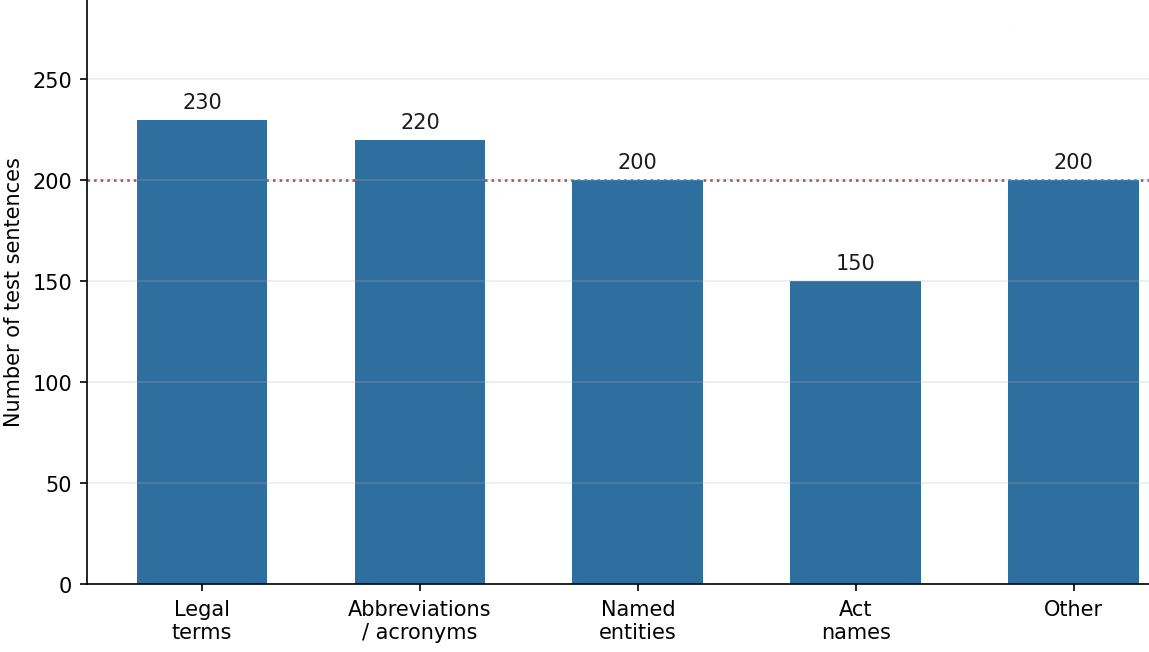}
    \caption{Sentence types in the \num{1000}-pair benchmark test set.}
    \label{fig:test_type}
\end{figure}

The benchmark test set contains \num{1000} sentence pairs, with \num{21990} English tokens and \num{23150} Mizo tokens. Instead of choosing these examples at random, we deliberately focus on the kinds of sentences that make legal translation difficult: those with legal terms (230 pairs), abbreviations and acronyms (220), named entities (200), names of statutes and acts (150), and general legal writing (200); see Figure~\ref{fig:test_type}. The relatively high type–token ratio (0.180 in English, 0.159 in Mizo) is mainly due to the smaller size of this split.

\begin{figure}[!t]
    \centering
    \includegraphics[width=0.90\linewidth]{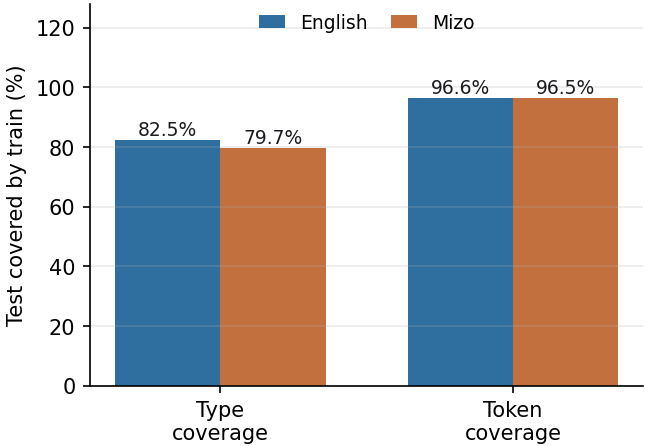}
    \caption{Test set coverage by training set vocabulary.}
    \label{fig:overlap}
\end{figure}

Figure~\ref{fig:overlap} shows how much the test vocabulary overlaps with the training data: \SI{82.5}{\percent} of English and \SI{79.7}{\percent} of Mizo distinct word types in the test set also appear in training, and coverage at the token level is \SI{96.6}{\percent} and \SI{96.5}{\percent}, respectively. This means that the test set is challenging in terms of unique words—about one in five Mizo word types is new—yet still manageable when reading continuous text.

\subsection{Model}
We use NLLB-200 \cite{nllb2022}, a Meta AI model that covers 200 languages, and trained on substantial mined parallel datasets. It is the natural choice here because it is one of the few large-scale open-source translation models with explicit support from Mizo (\texttt{lus\_Latn}), and serves as a strong baseline for low-resource translation. We evaluate both off the shelf and after fine-tuning and refer to the two systems as NLLB-200 \num{1.3}B and NLLB-FT.

\subsection{Fine-tuning and implementation}
Due to computational constraints, we use the \num{1.3} billion parameter variant of NLLB-200. Fine-tuning was performed with the Hugging Face \texttt{transformer} library \cite{wolf2020transformers} on a single NVIDIA A100 GPU, using AdamW \cite{loshchilov2019adamw} with a learning rate of $5\times10^{-5}$ and a linear warm-up schedule. We train for 5 epochs with a batch size of 4 and leave all other hyperparameters at their library defaults. We translate in the English$\rightarrow$Mizo direction throughout.

\subsection{Metrics}
We report three complementary automatic metrics.
\begin{itemize}
    \item \textbf{BLEU}: case-sensitive sacreBLEU \cite{post2018sacrebleu}, which fixes tokenisation and makes scores reproducible. Measures n-gram precision against the reference.
    \item \textbf{chrF++}: a character-level n-gram F-score \cite{popovic2017chrfpp}, well suited to a morphologically rich language such as Mizo and robust to tokenization differences.
    \item \textbf{TER}: translation edit rate \cite{snover2006ter}, the minimum number of insertions, deletions, substitutions, and shifts needed to convert the output to reference, normalized by reference length. Lower is better; it approximates the post-editing effort.
\end{itemize}

\section{Results}
\label{sec:results}

\begin{table}[t]
\centering
\small
\caption{Machine translation results on the \num{1000}-pair benchmark.}
\label{tab:mt_results}
\begin{tabular}{lccc}
\toprule
\textbf{System} & \textbf{BLEU $\uparrow$} & \textbf{chrF++ $\uparrow$} & \textbf{TER $\downarrow$} \\
\midrule
NLLB-200 \num{1.3}B & 20.31 & 42.56 & 77.81 \\
\textbf{NLLB-FT}    & \textbf{27.47} & \textbf{50.39} & \textbf{65.68} \\
\bottomrule
\end{tabular}
\end{table}

\begin{table}[t]
\centering
\small
\caption{NLLB-FT performance by test-set stratum.}
\label{tab:stratum-results}
\begin{tabular}{lrrrr}
\toprule
\textbf{Stratum} & \textbf{$n$} & \textbf{BLEU $\uparrow$} & \textbf{chrF++ $\uparrow$} & \textbf{TER $\downarrow$} \\
\midrule
Legal terms       & 230  & 19.71 & 46.43 & 69.70 \\
Abbreviations     & 220  & 34.05 & 53.58 & 63.73 \\
Named entities    & 200  & 26.33 & 53.08 & 62.42 \\
Act names         & 150  & 32.18 & 52.60 & 63.34 \\
Others     & 200  & 21.14 & 46.27 & 68.38 \\
\midrule
\textbf{Overall}  & \num{1000} & 27.47 & 50.39 & 65.68 \\
\bottomrule
\end{tabular}
\end{table}

Table~\ref{tab:mt_results} shows the overall results. When we fine-tune the model on legal texts from the target domain, all scores improve compared to the off-the-shelf model: BLEU goes from \num{20.31} to \num{27.47} ($+\num{7.16}$), chrF++ from \num{42.56} to \num{50.39} ($+\num{7.83}$), and TER drops from \num{77.81} to \num{65.68} ($-\num{12.13}$). Since we see gains in both a surface-based metric and an edit-based metric, the improvements are due to better wording and sentence structure, not quirks of a single measure. The size of this improvement also makes it clear that general-domain multilingual pre-training by itself is not sufficient for Mizo legal translation.

\subsection{Analysis by stratum}
\label{ssec:stratum}

Overall scores hide patterns that are specific to legal texts, so we take a closer look at NLLB-FT using the stratified benchmark. Throughout this analysis, we compare the system’s output to the \emph{actual choices made by the human translator} in the reference, instead of measuring it against some abstract idea of “correctness.” This approach helps us tell apart real mistakes from acceptable differences in how words or terms are borrowed.

Table~\ref{tab:stratum-results} shows that BLEU scores vary by more than \num{14} points across different strata. Scores are highest when the source text includes content that can be copied directly into the translation—such as abbreviations (\num{34.05}) and statute names (\num{32.18})—and lowest for legal terms (\num{19.71}) and general prose (\num{21.14}), where the model has to pick an appropriate Mizo word instead of reusing the English one. In other words, the system is better at \emph{keeping} English words than at \emph{replacing} them with Mizo, and the next section shows that this single pattern explains most of the errors we see.

\section{Error Analysis}
\label{sec:error-analysis}

To see where NLLB-FT still fails, we inspect its output on the benchmark. The ordering in Table~\ref{tab:stratum-results} is partly an artifact: strings that can be copied are copied, and copying is rewarded by surface metrics even when the resulting segment is not a translation. We therefore analyze the three critical strata separately, combining automatic extraction of term, entity, and acronym mentions with manual inspection of the flagged segments.

\subsection{Legal terminology}
\label{ssec:err-legalterms}

Section~\ref{ssec:borrowing} showed that professional translators borrow selectively. The dominant failure of NLLB-FT is that it does not reproduce this pattern: it keeps English words where the translator uses a Mizo one. Across the \num{230} legal-term segments the model matches the reference choice in \num{150} (\SI{65.2}{\percent}). Where the two differ, over-retention dominates: in \num{50} segments (\SI{21.7}{\percent}) the model copies the English term although the reference nativizes it, against only \num{19} segments (\SI{8.3}{\percent}) in the opposite direction (Table~\ref{tab:err-legalterms}). The system therefore borrows roughly three times more often than the translator. A smaller but more serious class is meaning-shifting substitution (\num{5} segments), where the model picks a Mizo word that changes the legal meaning, for example rendering \textit{accused} as \textit{pawikhawihtu} (`troublemaker').

\begin{table}[t]
\centering
\small
\caption{Legal-terminology errors in the NLLB-FT output on the
\num{230}-segment legal-term stratum. Percentages are of segments in
the stratum; $n$ counts affected segments.}
\label{tab:err-legalterms}
\begin{tabular}{lrl}
\toprule
\textbf{Error type} & \textbf{$n$ (\%)} & \textbf{Typical case} \\
\midrule
Over-retention of      & 50 (21.7) & \textit{learned}, \textit{evidence} \\
\quad English form     &           & kept verbatim \\
Spurious nativisation  & 19 (8.3)  & \textit{affidavit} $\rightarrow$ native \\
Variant equivalent     & 7 (3.0)   & \textit{order}: \textit{thupek} / \\
\quad (not an error)   &           & \textit{thuchhuah} \\
Meaning-shifting       & 5 (2.2)   & \textit{accused} $\rightarrow$ \\
\quad substitution     &           & \textit{pawikhawihtu} \\
Untranslated copy      & 4 (1.7)   & segment left in English \\
Truncation / omission  & 2 (0.9)   & clause dropped \\
\bottomrule
\end{tabular}
\end{table}

\subsection{Named entities}
\label{ssec:err-entities}

The model usually handles named entities well. Out of the \num{842} entity mentions that the reference version copies word for word, the model reproduces \num{814} of them exactly (\SI{96.7}{\percent}). This means that party names, place names, case numbers, and dates are mostly reliable (see Table~\ref{tab:err-entities}).

The \num{15} corrupted forms are systematic rather than random. Plural institutional names are silently singularized (\textit{District Councils} $\rightarrow$ \textit{District Council}), which can change the body to which it being referred to; statute titles are rewritten in canonical form (\textit{Criminal Procedure Code, 1973} $\rightarrow$ \textit{Code of Criminal Procedure, 1973}), which is bibliographically correct but no longer the form used by the court; and low-frequency names drift orthographically (\textit{Practices} $\rightarrow$ \textit{Practises}).

\begin{table}[t]
\centering
\small
\caption{Named-entity behaviour in the NLLB-FT output, over the
\num{842} entity mentions that the reference transfers verbatim.}
\label{tab:err-entities}
\begin{tabular}{lrl}
\toprule
\textbf{Behaviour} & \textbf{$n$ (\%)} & \textbf{Example} \\
\midrule
Transferred correctly & 814 (96.7) & \textit{Jogen Kalita} \\
Corrupted form        & 15 (1.8)   & \textit{Practices} $\rightarrow$ \\
                      &            & \textit{Practises} \\
Omitted               & 13 (1.5)   & entity dropped \\
\bottomrule
\end{tabular}

\centering
\small
\caption{Abbreviation and acronym behaviour in the NLLB-FT output, over
the \num{208} acronym mentions that the reference retains.}
\label{tab:err-abbrev}
\begin{tabular}{lrl}
\toprule
\textbf{Behaviour} & \textbf{$n$ (\%)} & \textbf{Example} \\
\midrule
Acronym preserved    & 203 (97.6) & \textit{ND\&PS}, \textit{IPC} \\
Lost or paraphrased  & 4 (1.9)    & \textit{P.W.1} $\rightarrow$ \\
                     &            & \textit{Thuhretu-1 na} \\
Case / punct.\ drift & 1 (0.5)    & \textit{Cr.PC} $\rightarrow$ \textit{Cr.Pc} \\
\bottomrule
\end{tabular}
\end{table}

\subsection{Abbreviations and acronyms}
\label{ssec:err-abbrev}

The acronym strings are the most reliable part of the output (Table~\ref{tab:err-abbrev}). Of the \num{208} acronym mentions that each reference includes, \num{203} (\SI{97.6}{\percent}) are copied exactly. Four are either dropped or loosely translated (for example, \textit{P.W.1} becomes \textit{Thuhretu-1 na}), and one has a minor change in the letter case (\textit{Cr.PC} $\rightarrow$ \textit{Cr.Pc}). Differences like \textit{ND\&PS} versus \textit{NDPS}, or \textit{Cr.P.C.} versus \textit{Cr.PC} and \textit{CrPC}, already appear in both the source and the reference, and the system preserves them as they are. This matches the broader corpus pattern, where legal abbreviations such as \textit{IPC} (\SI{100.0}{\percent}) and \textit{FIR} (\SI{100.0}{\percent}) are kept on the Mizo side near their original rates.

\section{Conclusion}
\label{sec:conclusion}

We introduce the first English–Mizo parallel corpus in the legal domain: \num{13168} sentence pairs taken from Indian court judgments. Professional native Mizo translators worked at the document level, and the sentence pairs were then manually aligned and independently validated. Two findings stand out. First, when Indian legal English is translated into Mizo, there is a clear pattern of borrowing: the names of institutions and laws are kept almost unchanged, while descriptions of events, processes, and outcomes are translated using native Mizo words. We interpret this pattern as a way to keep legal references very stable. Second, real-world Mizo writing does not use diacritics consistently. As a result, deciding whether and how to normalize diacritics is still an open preprocessing option, not a settled one.

Our baseline results show that fine-tuning on in-domain data really helps: the NLLB-200 \num{1.3}B model improves by $+\num{7.16}$ BLEU on a carefully sampled legal test set. However, our error analysis shows that the remaining problems are exactly in the areas where English borrowing is most likely. The fine-tuned system copies English words in the translation about three times more often than a human translator does.

This suggests two directions. On the modeling side, terminology-aware objectives or constrained decoding informed by a retention lexicon could help a system learn \emph{when} to borrow rather than borrowing by default. On the evaluation side, terminology-sensitive metrics are needed, since BLEU and chrF++ credit copying.


\section{Acknowledgements}
We thank the professional translators and annotators who built this resource.
The funding for data collection in this work was provided through the Mizo Language Model Training Hackathon, organized by the Digital India BHASHINI Division (Government of India) and the Government of Mizoram.

\section{Generative AI Use Disclosure}
Generative AI tools were used only for the language editing and polishing of
the manuscript. They were not used to produce any part of the research
content, the corpus, or the experiments.

\bibliographystyle{IEEEtran}
\bibliography{mybib}

\end{document}